\documentclass[letterpaper]{article} 
\usepackage{aaai23}  
\usepackage{times}  
\usepackage{helvet}  
\usepackage{courier}  
\usepackage[hyphens]{url}  
\usepackage{graphicx} 
\usepackage{natbib}  
\usepackage{caption} 
\usepackage{algorithm}
\usepackage{algorithmic}
\usepackage{makecell}
\usepackage{newfloat}
\usepackage{listings}
\DeclareCaptionStyle{ruled}{labelfont=normalfont,labelsep=colon,strut=off} 
\floatstyle{ruled}
\newfloat{listing}{tb}{lst}{}
\floatname{listing}{Listing}
\usepackage{amsmath,amssymb} 
\usepackage{booktabs}
\usepackage{eucal}
\usepackage{bm}
\usepackage{multirow}
\usepackage{subfigure}
\def\network{ANA-Net}
\newtheorem{proposition}{Proposition}
\usepackage[capitalize]{cleveref}
\crefname{section}{Sec.}{Secs.}
\Crefname{section}{Section}{Sections}
\Crefname{table}{Table}{Tables}
\crefname{table}{Table}{Tables}

\title{Learning Second-Order Attentive Context for Efficient Correspondence Pruning}
\author {
    Xinyi Ye, 
    Weiyue Zhao, 
    Hao Lu, 
    Zhiguo Cao$^{\ast}$
}
\affiliations {
    Key Laboratory of Image Processing and Intelligent Control, Ministry of Education School of \\ Artificial Intelligence and Automation, Huazhong University of Science and Technology\\
    $\{$xinyiye,zhaoweiyue,hlu,zgcao$\}$@hust.edu.cn
}

\begin{document}

\maketitle

\begin{abstract}
Correspondence pruning aims to search consistent correspondences (inliers) from a set of putative correspondences. 
It is challenging because of the disorganized spatial distribution of numerous outliers, especially when putative correspondences are largely dominated by outliers. 
It is more challenging to ensure effectiveness while maintaining efficiency. In this paper, we propose an effective and efficient method for correspondence pruning.
Inspired by the success of attentive context in correspondence problems, we first extend the attentive context to the first-order attentive context and then introduce the idea of attention in attention (ANA) to model second-order attentive context for correspondence pruning.
Compared with first-order attention that focuses on feature-consistent context, second-order attention dedicates to attention weights itself and provides an additional source to encode consistent context from the attention map.
For efficiency, we derive two approximate formulations for the naive implementation of second-order attention to optimize the cubic complexity to linear complexity, such that second-order attention can be used with negligible computational overheads. We further implement our formulations in a second-order context layer and then incorporate the layer in an ANA block.
Extensive experiments demonstrate that our method is effective and efficient in pruning outliers, especially in high-outlier-ratio cases. Compared with the state-of-the-art correspondence pruning approach LMCNet, our method  runs $14$ times faster while maintaining a competitive accuracy.
\end{abstract}

\section{Introduction}
\label{sec:intro}
\begin{figure}[t]
    \centering
    \includegraphics[width=0.45\textwidth]{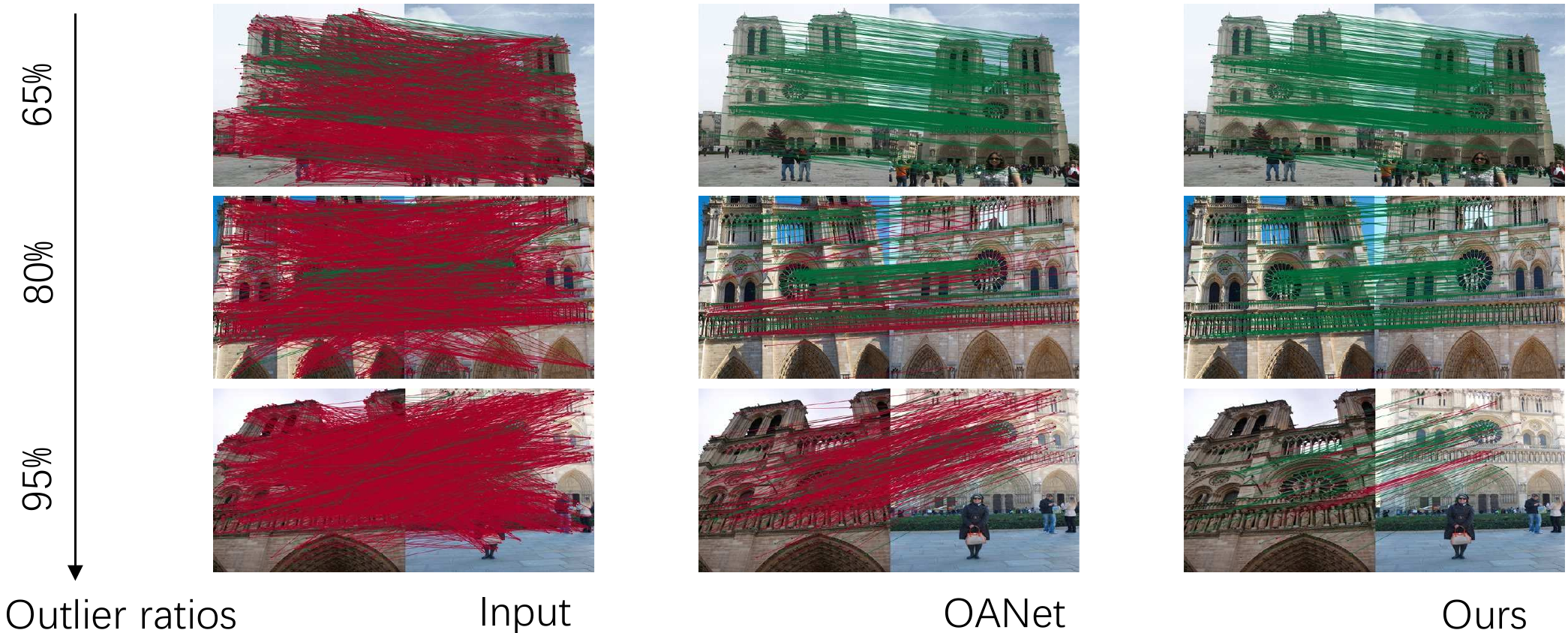}
    \caption{Imbalanced putative correspondences. Given a set of correspondences ($1^\text{st}$ column) with increased outlier ratios (from top to bottom, $65\%$, $80\%$, and $95\%$), OANet loses efficacy gradually and fails to reject mass outliers(red), while our method keeps($3^\text{rd}$ column) a high accuracy.}
    \label{fig:fig1}
\end{figure}

Finding correspondences is a fundamental problem in many computer vision tasks, such as structure-from-motion~\cite{sfm}, inpainting~\cite{zhou2021transfill}, and simultaneous location and mapping~\cite{slam}. However, correspondences searched with existing detector-based descriptors~\cite{superpoint,hardnet,sift,orb,lfnet} abound with outliers (\cref{fig:fig1},$1^{\text{st}}$ column). To prevent outliers from disturbing downstream tasks, correspondence pruning (or consistency filtering) is by default used as a postprocessing procedure  to identify consistent correspondences (inliers)~\cite{gms,ma2019locality,cne}.

Recently, deep models~\cite{cne,acne,oanet} are proposed as a plug-in to prune outliers and report promising results. Unfortunately, as outlier ratios increase, they lose efficacy~\cite{lmcnet,zhao2021progressive} and even identify outliers as inliers (\cref{fig:fig1}, $2^\text{nd}$ column). Liu \textit{et al.}~\cite{lmcnet} propose LMCNet that leverages the motion coherence to distinguish correspondences with discriminative eigenvectors. However, when the singular value decomposition deals with a large matrix, it can be computationally expensive: For correspondence pruning, the size of a matrix used to model point-to-point relation is often larger than $10^3\times10^3$ and thus leads to one-order-of-magnitude extra inference time. This may be problematic when facing practical applications, especially the real-time ones. Here we explore how to tackle the correspondence pruning problem both effectively and efficiently.

In correspondence pruning, a fundamental problem is \textit{how to distinguish inliers from outliers}. 
Previous works~\cite{oanet,acne} believe inliers are consistent in both local and global context so that local-global context is encoded to help to distinguish inliers from outliers. However, Zhao \textit{et al.}~\cite{nmnet} points out local context would be unstable due to randomly distributed outliers. Global context suffers from the same affliction because inliers would be submerged in massive outliers. This results in the problem of inconsistent context. The similar inputs are more likely to lead to similar outputs and it is perhaps one of the reasons why these methods lose their efficacy when massive outliers occur. 
A question of interest is \textit{how to efficiently encode consistent context}. 

ACNe~\cite{acne} suggests that the attention mechanism can guide a network to focus on a subset of inlier features and thus can suppress outliers. ACNe uses this mechanism to generate the so-called attentive context that is consistent for the inliers.
Following this spirit, we delve deep into attentive context and observe that the naive attentive context modeled by feature-based similarity, which could be interpreted as first-order attention, is not sufficient. Instead we find taking additional second-order attention as a supplementary is helpful. This finding is based on an observation that, inliers share similar attention weights, while outliers do not. We therefore introduce the idea of attention in attention to model second-order attentive context. The key idea is to introduce extra distinctive information from attention weights such that inliers can be grouped and distinguished.

To this end, we propose to jointly model first- and second-order attentive context, featured by feature- and attention-consistent context, respectively. In particular, we first encode with a first-order attention map generated by self-attention~\cite{attention}. We then seek attention-consistent context from a second-order similarity matrix generated by attention weights. For efficient end-to-end training, we discuss the most effective part of the naive implementation of second-order attention, optimize the cubic-complexity implementation to linear complexity, and propose a Second-Order Context (SOC) layer. By integrating the SOC layer into a permutation-invariant block, we further present an iterative Attention in Attention Network (ANA-Net) that can effectively address correspondence pruning. We show that this network is effective in pruning outliers, especially in high-outlier-ratio cases (\cref{fig:fig1}, $3^\text{rd}$ column).

We conduct extensive experiments to demonstrate the effectiveness and generalization of our approach on camera pose estimation and correspondence pruning. Thanks to the consistent context, \network{} achieves superior performance over most baselines and shows its potentials to address multi consistency problems. Meanwhile, \network{} maintains comparable performance against LMCNet while running $14$ times faster than it. 

\noindent\textbf{Contributions.}We introduce the idea of attention in attention that essentially models similarity between attention weights, which can also be interpreted as second-order attention. Technically, we present a formulation that models such second-order information and further derive two approximated formulations to reduce computational complexity. We show that our formulations can be implemented as a network layer and can be used to cooperate the first-order attention to address the problem of correspondence pruning with high accuracy, low computational cost and robust generalization.

\section{Related Work}
\noindent\textbf{Deep Correspondence Pruning.} RANSAC~\cite{ransac} and its variants~\cite{chum2005two,adalam,magsac,barath2018graph,barath2020magsac++} always suffers from high-outlier-ratio cases, thus deep filters are critical to reject most outliers and guarantee a more manageable correspondence set for RANSAC. And pioneering works such as DSAC~\cite{brachmann2017dsac} and CNe~\cite{cne} prove the superiority of neural networks for pruning correspondences. Since sparse correspondences are unordered and irregular, 
convolution operators~\cite{Krizhevsky2012ImageNetCW} cannot be adopted directly. Inspired by GNN~\cite{scarselli2008graph}, OANet~\cite{oanet} proposes the generalized differentiable pooling~\cite{ying2018hierarchical} and unpooling to cluster correspondences. To improve the robustness of CNe~\cite{cne}, ACNe~\cite{acne} introduces the attention mechanism to focus the normalization of the feature maps of inliers. According to the affine attributes, NM-Net~\cite{nmnet} mines compatibility-specific neighbors to aggregate features with a compatible Point-Net++~\cite{pointnet++} architecture.
Despite achieving satisfactory performance in correspondence pruning, they can still suffer from high-outlier-ratio cases.

\noindent\textbf{Consensus in Correspondences.}
One cannot classify an isolated correspondence as inlier or outlier~\cite{zhao2021progressive}. Consensus is the motivation for algorithms to distinguish outliers. There are generally two types of classical consensus in correspondences: global consensus and local consensus. RANSAC-related work is driven by the former consensus and identifies correct correspondences by judging whether correspondences conform to a task-specific geometric model such as epipolar geometry or homography~\cite{lmcnet}. The latter is more intuitive. For instance, LPM~\cite{ma2019locality} assumes local geometry around correct correspondences does not change freely. CFM~\cite{chen2015co} believes local structural or local textural variations estimated by a consistent pair of transformations should be similar. In this paper, the feature-consistent context generated with the first-order attention is an application of both types of consensus, while the attention-consistent context captured with second-order attention does not belong to either of them. It is dedicated to attention itself and we view it as a representation of attention consensus.

\noindent\textbf{Self-Attention Mechanism.}
Inspired by the success of self-attention mechanism in NLP\cite{bahdanau2014neural,attention,kim2017structured}, many researchers notice its ability to focus on input of interest and to capture long-range context. Wang \textit{et al.}~\cite{wang2018non} combines CNN-like architectures with self-attention and achieves non-local feature representation. Pan \textit{et al.}~\cite{pan2021transview} builds the cross context of view boundaries via the attention mechanism in image cropping. Wang \textit{et al.}~\cite{wang2022interior} models pixel-level attention for weak small object detection.
Carion \textit{et al.}~\cite{carion2020end} and Misra \textit{et al.}~\cite{transformerFor3dDt} adopt the attention mechanism to reason about the relation of objects in 2D and 3D detection.
For correspondence pruning, ACNe
~\cite{acne} focuses on feature normalization w.r.t.\ a subset of inliers. Due to the non-local property of the self-attention mechanism, in this work we validate its use to deal with high-outlier-ratio cases in correspondence pruning. 

\section{Learning Consistent Context}
\subsection{Problem Formulation}
Given an image pair $(\bm{I}^s, \bm{I}^t)$, a putative correspondence set  $\bm{C} \in \mathbb{R}^{N \times 4}$ can be established via nearest neighbor feature matching. This set is our input to the correspondence pruning method. The goal of correspondence pruning is to predict a weighted vector  $\bm{w}=[w_1,...,w_N] \in \mathbb{R}^{N \times 1}$, where $w_i \in [0,1)$ indicates the inlier probability of $i^\text{th}$ correspondence. The input correspondence is conventionally a concatenated $4$D vector representing pairwise keypoints coordinates. However, the disorganized spatial distribution of numerous outliers makes it difficult to directly filter unreliable correspondences with a classical method such as RANSAC. To distinguish between inliers and outliers in the high-dimensional feature space, a few learnable methods~\cite{cne,lmcnet,nmnet,oanet} adopt an iterative strategy with multiple MLP layers to capture contextual information for better outlier rejection.

\subsection{First-Order Attention} \label{Sec:first-order}
\begin{figure}[t]
  \centering
    \includegraphics[width=0.43\textwidth]{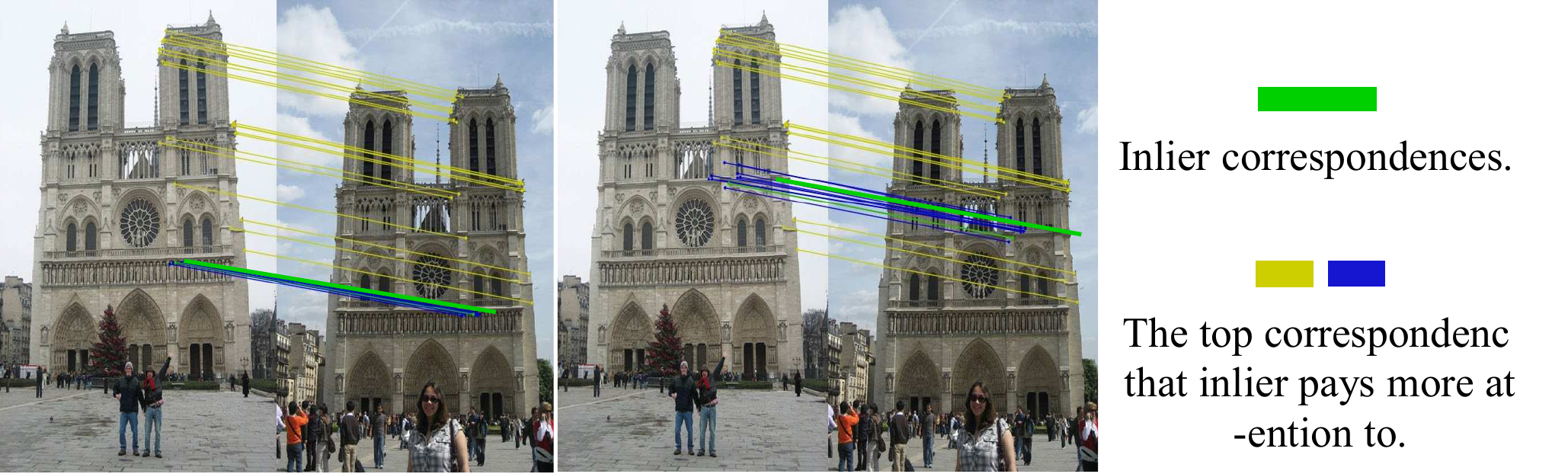}
    \caption{Motion coherence of inliers indicated by first-order attention. We select two inliers green (Left) and (Right) in the same image pair. To visualize the top correspondences that they care about, we color them according to the attention weights respectively. Specifically, yellow denotes the mutual correspondences shared by different inliers, and the rest are shown in blue. It suggests inliers pay more attention to each other with higher attention weights.}
  \label{fig:Attention distribution}
\end{figure}

Compared with the MLP-based techniques that integrate consistency information with restricted neighbor correspondences, 
ACNe~\cite{acne} attempts to enlarge the receptive field with the attention mechanism. The attention used by ACNe can be thought of as first-order attention, because it essentially models the similarity between feature maps. By normalizing the feature map with an attentive weight function, ACNe could effectively cluster features and thus separate inliers from outliers. Inspired by~\cite{acne}, we consider optimizing the feature map for better classification by fusing attentive context. As shown in Fig.~\ref{fig:Attention distribution}, each inlier shares the same motion coherence with certain correspondences. In other word, correspondences with similar motion trends are likely to focus on each other. Therefore, we instead use self-attention to guide features to encode consistent context for inliers. It is worth noting that, self-attention is also first-order attention, despite the fact that it is represented by a 2D map. Next we show how to encode correspondence features to obtain consistent context
with a first-order attention map. 

Given a first-order attention map $\bm{A}=[\bm{a}_1,...,\bm{a}_N]\in \mathbb{R}^{N \times N}$, where $\sum_{j}a_{ij}=1$ denotes the feature similarity between the $i^\text{th}$ correspondence and the correspondence set, we can acquire the feature-consistent context $\bm{v}_i$ by $\bm{v}_i=\sum_{j}a_{ij}\bm{x}_j\,$ where $\bm{x}_j$ is the feature vector of the $j^\text{th}$ correspondence. Essentially, feature-consistent context exploits the motion-consistent nature of inliers to filter outliers that have irregular motion distributions. Considering that the inliers have similar feature patterns, a question of interest is whether inliers possess the latent similarity of attention weights. To delve attention-consistent context among correspondences, we introduce second-order attention.

\subsection{Second-Order Attention}  
\label{sec:ASC}
From Fig.~\ref{fig:Attention distribution}, we notice that different inliers share mutual attentive correspondences (yellow), which suggests inliers may also share similar attention patterns. A key idea of our method is to explore attention-consistent context with a second-order attention map as an explicit measurement of attentive consistency.

Given the first-order attention map $A$, we obtain the second-order attention map by $\bm{W}=\bm{A}^{T}\bm{A} \in \mathbb{R}^{N \times N}$, where $w_{ij}=\sum_{k}a_{ki}a_{kj}$ indicates the similarity of attention weights between the $i^\text{th}$ and $j^\text{th}$ correspondences.  Our goal is to extract discriminative features from $\bm{W}$ as the attention-consistent context to separate the inliers from outliers. Since the matrix $\bm{W}$ shares the same spirit of spectral clustering, a natural idea is to follow the formulation of spectral clustering to extract features. In general, spectral clustering defines a degree matrix $\bm{D}=\text{ diag}([d_i=\sum_{j}w_{ij}])$ of $\bm{W}$ and a Laplacian matrix $\bm{L}=\bm{D}-\bm{W}$, to derive classifiable eigenvectors by applying singular value decomposition~(SVD) on $\bm{L}$. Due to the fact that spectral clustering is based on the theory of graph cuts, it cannot be directly applied to our problem since the given inconsistent motions among outliers. A huge time consumption is introduced as well because of the large matrix decomposition, which deviates from our goal. Yet , it still inspires us to extract discriminative context encoded in $\bm{L}$ via a differentiable operation.

As described in OANet~\cite{oanet}, the operation ought to be permutation-invariant such that the context is permutation-invariant, which is critical for correspondence pruning. Inspired by Diff-pool in OANet~\cite{oanet}, where input features are mapped to clusters, we utilize inner product like Diff-pool to map $\bm{L} \in \mathbb{R}^{N \times N}$ into a feature map with fixed dimension to be embeded into the network by the following proposition.

\begin{proposition}[Cubic Form.] Let $\bm{l}_i$ denote the $i^\text{th}$ column vector of $\bm{L}$. The distinctive information of the $i^\text{th}$ correspondence is encoded by 
   \begin{equation} 
       h_i=\sqrt{\bm{l}_i^T\bm{l}_i}=\sqrt{(\sum_{j,j\neq i}w_{ij})^2+\sum_{j,j\neq i}w_{ij}^2}\,,
       \label{eq:Cubic Form}
   \end{equation}
   where $h_i$ is called the attention-consistent context. $h_i$ requires $N^2+N+1$ multiplications to be computed from the first-order attention map. For $N$ correspondences, the computational complexity~(ignoring the impact of summation) of the cubic form is $\mathcal{O}(N^3)$.
\end{proposition}

Essentially, Eq.~\eqref{eq:Cubic Form} 
executes a cumulative operation on elements of $\bm{W}$ except the self-correlation $w_{ii}$. There is an understanding derived from~\cref{fig:Attention distribution}: inliers have more $w$'s of large values while outliers do not, which suggests the values of Eq.~\eqref{eq:Cubic Form} for inliers  will be different from outliers. In fact, Eq.~\eqref{eq:Cubic Form} achieves the same thing like the SVD operation in spectral clustering: extracting classifiable features for inputs to categorize. With the priors form the observation in~\cref{fig:Attention distribution}, it performs with relatively fewer computational overheads. However, the computational overheads of it are still heavy for correspondence pruning because of the cubic complexity. A question remains:\textit{Is it possible to simplify the form while retaining distinguishability}. The answer is yes. Assume that $y_i=1$ in the Cauchy's Inequality $ (\sum_{i}x_i^2)(\sum_{i}y_i^2)\geq(\sum_{i}x_iy_i)^2$, we have
\begin{equation}
    N\sum_{i}x_i^2\geq(\sum_{i}x_i)^2\,.
    \label{eq:y=1}
\end{equation}
Meanwhile, since $w_{ij}>0$, we have $(\sum_{j,j\neq i}w_{ij})^2\geq\sum_{j,j\neq i}w_{ij}^2$. By inserting them into \cref{eq:y=1}, one can derive the upper and lower bound of \cref{eq:Cubic Form} as
\begin{equation}
    \sqrt{2}\sum_{j,j\neq i}w_{ij}\geq{h}_i\geq\sqrt{\frac{N+1}{N}}\sum_{j,j\neq i}w_{ij}\,.
    \label{eq:boundary}
\end{equation}
In this work, we use the upper bound as an approximation and obtain the quadratic formulation.
\begin{proposition}[Quadratic Form.] Let $w_{ij}$ denote the similarity of attention weights between the $i^\text{th}$ and $j^\text{th}$ correspondences such that $w_{ij}=\sum_{k}a_{ki}a_{kj}$. Eq.~\eqref{eq:Cubic Form} has a quadratic form, defined by
   \begin{equation}
       {h}_i=\sqrt{2}\sum_{j,j\neq i}w_{ij}=\sqrt{2}(\sum_{k}a_{ki}-\sum_{k}a_{ki}^2)\,.
       \label{eq:Quadratic Form}
   \end{equation}
   Since $h_i$ requires $N+1$ multiplications to compute, the computational complexity of the quadratic form 
   for all correspondences is $\mathcal{O}(N^2)$. 
\end{proposition}

While the quadratic form has significantly reduced the computational complexity from $\mathcal{O}(N^3)$ to $\mathcal{O}(N^2)$, we can further approximate the quadratic form to a linear form.
\begin{proposition}[Linear Form.] Let the upper bound of \cref{eq:Quadratic Form} be used for a second time according to \cref{eq:y=1}. A linear approximation of \cref{eq:Quadratic Form} takes the form
   \begin{equation}
        {h}_i=\sqrt{2}(\sum_{k}a_{ki}-\frac{1}{N}(\sum_{k}a_{ki})^2)\,.
    \label{eq:Linear Form}
   \end{equation}
   The computational complexity of the linear form is $\mathcal{O}(N)$.
\end{proposition}

We can expand $\sum_{k}a_{ki}$ as: $\sum_{k}a_{ki}=\sum_{k}(a_{ki}*1)=\sum_{k}\sum_{j}a_{ki}a_{kj}=\sum_{j}\sum_{k}a_{ki}a_{kj}=\sum_{j}w_{ij}=w_{ii}+\sum_{j,j\neq i}w_{ij}$, where $\sum_{j,j\neq i}w_{ij}$ is the first term of Eq.~(1), and $\sum_{k}a_{ki}$ is the first term of Eqs.~(4)\&(5). Despite the non-tight upper bound, the distinguishability of Eqs.~\eqref{eq:Quadratic Form}\&~\eqref{eq:Linear Form} is still reflected in the numerical differences like Eq.~\eqref{eq:Cubic Form}. They are thus suitable for encoding second-order attentive context without much accuracy loss. For end-to-end training, we present a Second-Order Context (SOC) layer to implement the formulations above.

\subsection{Attention in Attention Block}
\label{sec:anab}
\begin{figure}[t]
    \centering
    \subfigure[]{
    \includegraphics[width=0.40\textwidth]{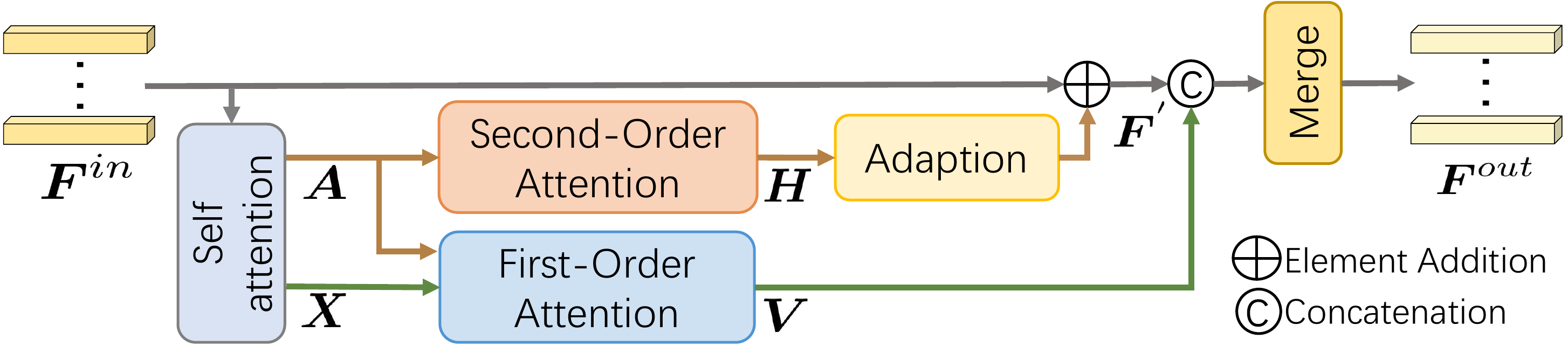}
    \label{fig:block-a}
    }
    \subfigure[]{
    \includegraphics[width=0.18\textwidth]{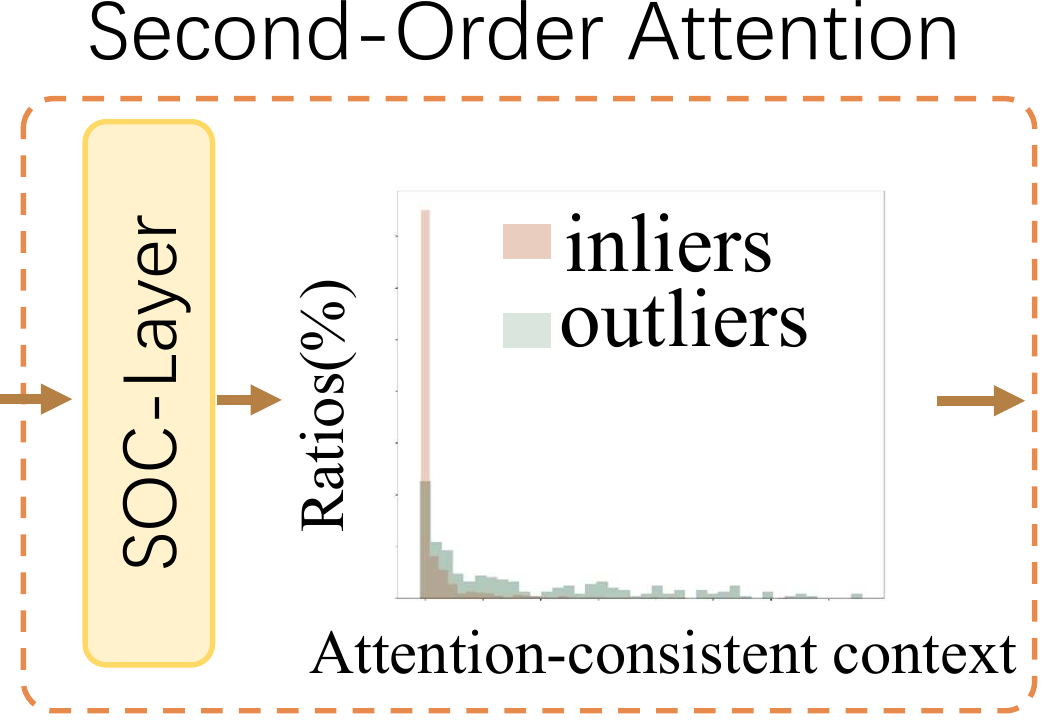}
    \label{fig:block-b}
    }
    \subfigure[]{
    \includegraphics[width=0.22\textwidth]{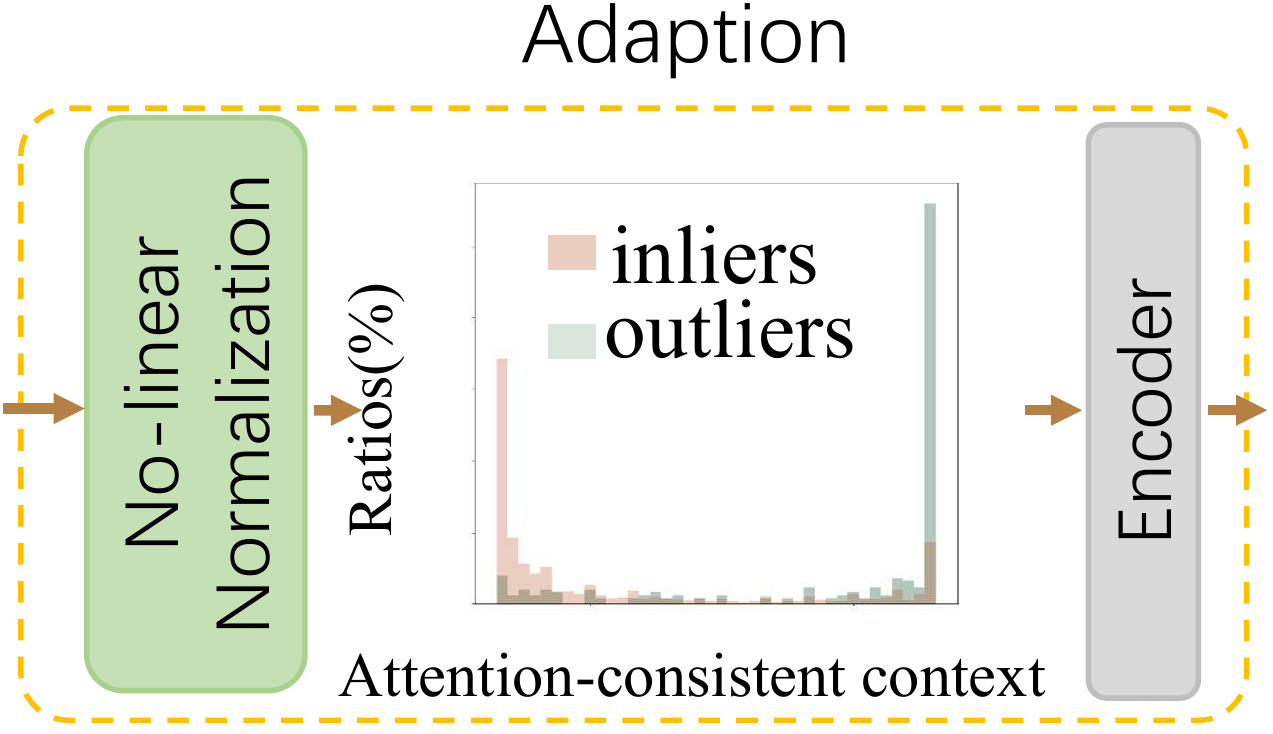}
    \label{fig:block-c}
    }
   
    \caption{
    Attention in attention block (ANA-Block). (a) is the overview of ANA-Block. It includes two components: attention-consistent context enhancement and feature-consistent context enhancement. (b) illustrates the process to extract attention-consistent context with second-order attention. (c) shows how the attention-consistent context is embedded into the ANA-Block.}
    \label{fig:block}
    
\end{figure}
\begin{figure}[!t]
\centering
    \includegraphics[width=0.40\textwidth]{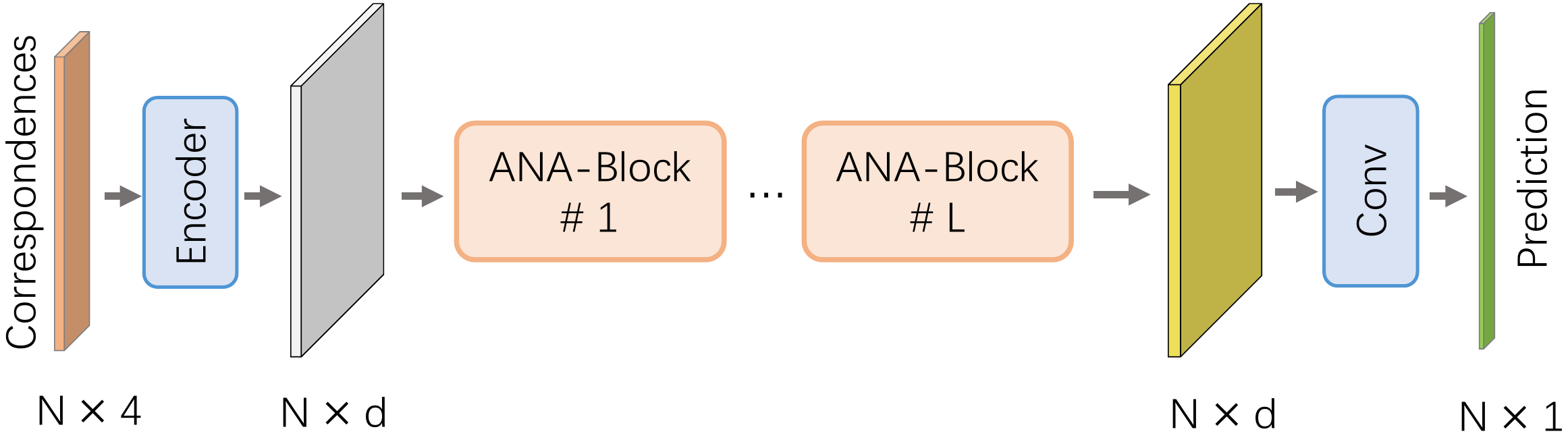}
  \caption{Attention in Attention Network (ANA-Net). $N$ represents the number of putative correspondences and $4$ denotes the concatenated $4$D vector a pair of keypoints coordinates in correspondences. }
  \label{fig:pipeline}
\end{figure}
Here we present the Attention in Attention Block (ANA-Block), aiming to encode both feature- and attention-consistent context. The feature-consistent context can be obtained from any off-the-shelf first-order attention,
 and the attention-consistent context is extracted by a SOC layer implementing the second-order attention formulations.
 The overview is shown in \cref{fig:block}(a).

Given a feature map $\bm{F}^{in}=[\bm{f}^{in}_1,...,\bm{f}^{in}_N]\in \mathbb{R}^{N\times d}$, the feature-consistent context $\bm{V}=[\bm{v}_1,...,\bm{v}_N]\in \mathbb{R}^{N\times d}$ and the attention-consistent context $\bm{H}=[h_1,...,h_N]\in \mathbb{R}^{N\times 1}$ can be obtained with the first-order and second-order attention, respectively. In particular, $\bm{H}$ is obtained by the SOC layer, which defines a mapping function $f_{SOC}$ such that $\bm{H}=f_{SOC}(\bm{A})$, where $\bm{A}$ is the first-order attention map. $f_{SOC}$ can be implemented by \cref{eq:Cubic Form}, \cref{eq:Quadratic Form}, or \cref{eq:Linear Form}. Since the attention-consistent context $\bm{H}$ has different characteristics from the initial feature map, we choose to embed it into the feature map rather than concatenating it. In this way, the embedded feature map encodes a hybrid representation $\bm{F}^{out}=[\bm{f}^{out}_1,...,\bm{f}^{out}_N]\in \mathbb{R}^{N\times d}$ for correspondence pruning, which is given by
\begin{equation}
        \bm{f}_i^{out}={\tt merge}([(\bm{f}_i^{in}+\psi(h_i))||\phi(\bm{v}_i)])\,,
\end{equation}
where $[\cdot||\cdot]$ denotes the concatenation operator, $\psi(\cdot)$ is a preprocessing step (discussed below) before embedding attention-consistent context into an ANA-Block, $\phi(\cdot)$ denotes a fusion function for feature-consistent context such as a shared MLP, $\tt merge(\cdot)$ is also a MLP that aggregates context and reduces the dimension from $2d$ to $d$. Next we discuss this process in detail, which is featured by attention-consistent context enhancement and feature-consistent context enhancement, as shown in \cref{fig:block}.
\vspace{5pt}

\begin{table}\Huge
\begin{center}
\resizebox{0.82\linewidth}{!}{
\begin{tabular}{@{}llccc@{}}
\toprule
Dataset & Method & AUC@${5^{\circ}}$ & AUC@${10^{\circ}}$ & AUC@${20^{\circ}}$ \\
\noalign{\smallskip}
\hline
\hline
\noalign{\smallskip}
\multirow{11}{*}{YFCC.} & {Magsac} & 28.24 & 44.86 & 61.53\\
		& {LPM} & 10.48 &18.91  & 29.26\\
		& {GMS}                   & 19.05 &32.35  & 46.79\\
		& {CODE}                   & 16.99 &30.23  & 43.85\\
    &LMF       &16.59  &29.14 &43.41\\
    &CNe     &25.18  &41.52 &57.36\\
    &ACNe      &28.90  &47.15 &63.95\\
    &OANet       &29.58  &47.71 &64.09\\
    &{LMCNet}  &\textbf{33.81}  &\textbf{53.10} &\textbf{69.97}\\
    &\network{}       &\underline{33.11} &\underline{52.02} &{68.77}\\
\hline
\multirow{9}{*}{SUN.}
    & {LPM} & 2.81 & 7.4  & 15.36\\
	& {GMS}            & 4.36  & 11.08  & 21.68\\
	& {CODE}    & 3.52 & 8.91  & 18.32\\
    &LMF &3.39  &8.91  &18.14\\
    &CNe   &5.17  &13.18 &25.74\\
    &ACNe      &4.80  &12.26 &23.75 \\
    &OANet       &4.81  &12.52 &24.52\\
    &{LMCNet} &\underline{5.78}  &\underline{14.61} &\underline{27.94}\\
    &\network{}          &\textbf{5.98}  &\textbf{14.95} &\textbf{28.35}\\
\bottomrule
\end{tabular}
}
\end{center}
\caption{Camera pose evaluation on YFCC100M and SUN3D. The best performance is in \textbf{bold}, and the second best is \underline{underlined}.}
\label{tab:sift}
\end{table}

\begin{table}[t]\Huge
  \centering
  \resizebox{0.90\linewidth}{!}{
  \begin{tabular}{@{}lcccccc@{}}
    \toprule
    Method & CNe & ACNe & OANet & LMCNet & \multicolumn{1}{l|}{\network{}} & Ref.\\
    \midrule
    runtimes  & 6.28  &11.52 &17.30& 212.99 &\multicolumn{1}{c|}{14.36} &  159.68
    \\
    
    \bottomrule
  \end{tabular}
    }
    \caption{Inference time~(ms). Ref. indicates SIFT$\&$NN.}
    \label{tab:parm}
\end{table}
\setlength{\tabcolsep}{1.4pt}

\noindent\textbf{Attention-Consistent Context Enhancement.} Here we explain $\psi(\cdot)$ in detail. As shown in \cref{fig:block-b}, we use a histogram to visualize attention-consistent context distribution for inliers and outliers, respectively. Since the distribution of inliers is dispersed, we design a non-linear normalization to cluster inliers when $h_{i}\rightarrow +\infty$ using an $\alpha$-parameterized ${\tt sigmoid}$ function such that 
\begin{equation}
     {\tt sigmoid}_\alpha(h_{i})=\frac{1}{1+\exp{(-\alpha h_{i}})}\,,
\end{equation}
where the learnable parameter $\alpha$ is used to adjust the level of non-linearity. Moreover, since the dimensionality of normalized $h_{i}$ does not equal to the original feature, we further map the normalized $h_{i}$ through an encoder to match the dimensionality. The enhanced feature $\bm{f}_{i}^{'} \in \mathbb{R}^{N\times d}$ is given by
\begin{equation}
    \bm{f}_{i}^{'}=\bm{f}_{i}^{in}+\psi(h_i)=\bm{f}_{i}^{in}+{\tt encoder}({\tt sigmoid}_{\alpha}(h_i))\,.
\end{equation}

\noindent\textbf{Feature-Consistent Context Enhancement.} Here we embodies $\phi(\cdot)$. To further integrate the feature-consistent context with $\bm{f}_{i}^{'}$, we choose a 
MLP to implement $\phi$ such that
\begin{equation}
    \bm{f}_{i}^{out}={\tt merge}([\bm{f}_{i}^{'}||{\rm MLP}(\bm{v}_i)])\,.
\end{equation}

Meanwhile, \cite{acne,oanet} emphasize that the permutation-invariance plays a crucial role in tasks with unordered data, such as correspondence pruning. It should be noted that our ANA-Block designed with consistent context is permutation-invariant. The proof of permutation-invariance can be found in the supplementary.

\subsection{Implementation Details}
In our implementation, we use the linear-form formulation for the SOC layer, and a multi-head structure in self attention\cite{attention} with $4$ heads is used in ANA-Blocks to generate the first-order attention map. For feature dimensionality, we follow a common setting $d=128$ for intermediate feature maps. Besides, we apply an iterative strategy to construct the framework by stacking L$=5$ ANA-Blocks (\cref{fig:pipeline}). The correspondence set $\bm{C}$ is normalized with camera intrinsic matrices if available, or is normalized into $[-1,1]$ according to the input image size in both training and testing. 

\noindent\textbf{Loss Function.} We use a weighted binary cross-entropy loss to balance positive and negative samples following~\cite{cne}, which takes the form
\begin{equation}
 \mathcal{L}(\bm{C})=\frac{1}{N}\sum_{i=1}^{N}\tau_{i}{\tt H}(y_i,{\tt S}(o_i))\,,
\end{equation}
where $o_i$ is the output of the $i^{th}$ correspondence, $y_i$ denotes the corresponding ground-truth label according to the epipolar distance and the threshold $\sigma=10^{-4}$, $\tt S(\cdot)$ represents the logistic function used in the binary cross entropy $\tt H(\cdot)$, and $\tau_i$ is a weight that balances the relative importance between positive and negative samples.

\begin{figure}[t]\centering
    \includegraphics[width=0.40\textwidth]{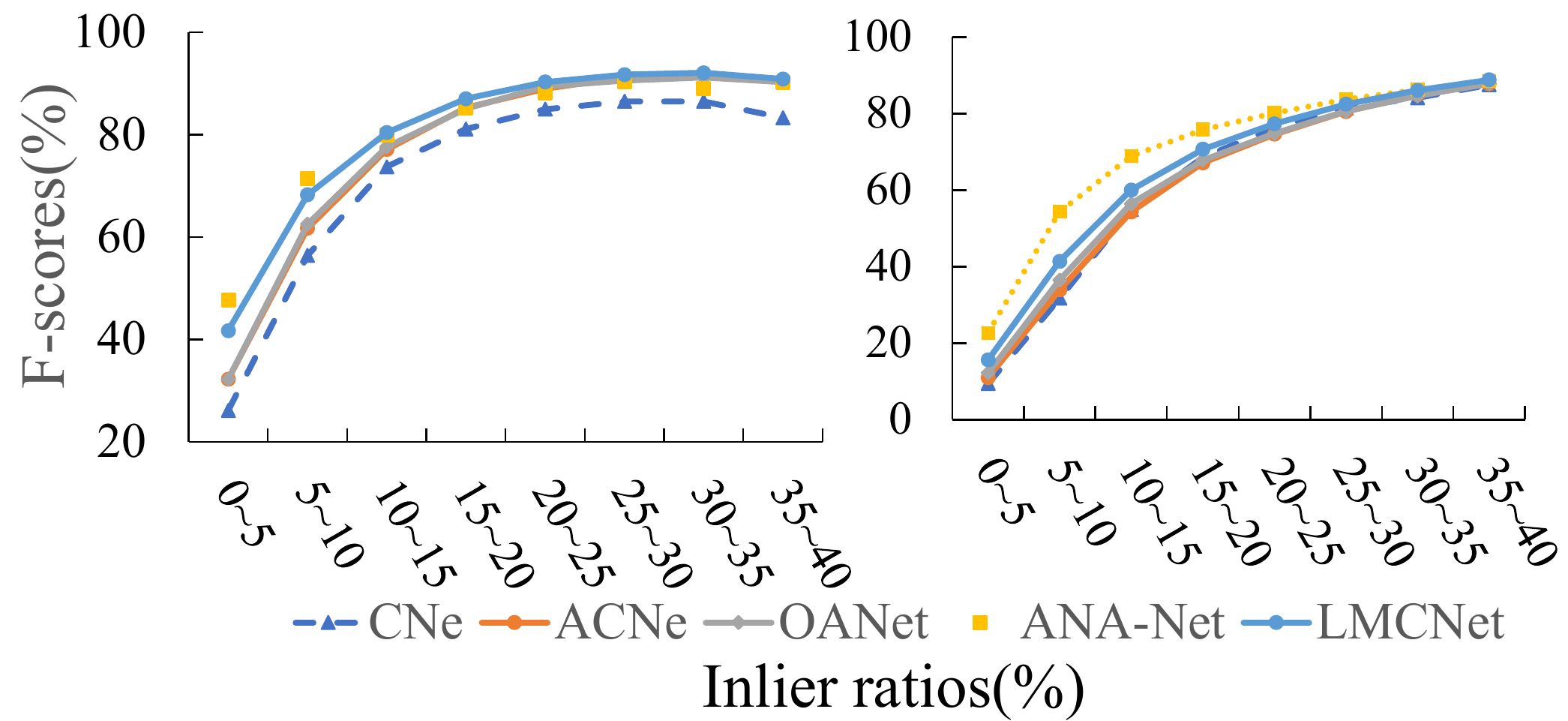}
  \caption{Performances with various inlier ratios. For different methods, we calculate F1-scores with various inlier ratios on both YFCC100M (left) and SUN3D (right).}
  \label{fig:short}
\end{figure}
 
\section{Experiments}
\subsection{Datasets and Evaluation Protocols}
\label{sec:protocols}
\noindent\textbf{Datasets.} The outdoor YFCC100M~\cite{yfcc} dataset and the indoor SUN3D~\cite{sun3d} dataset are used in camera pose evaluation. We apply the same data split following OANet~\cite{oanet}. The putative input correspondences are generated via nearest features matching of SIFT~\cite{sift} and SuperPoint~\cite{superpoint} with the maximum number of keypoints of $2000$ and $1024$, respectively. We consider correspondences with a small symmetric epipolar distance ($<10^{-4}$) as correct correspondences (inliers). An essential matrix is estimated with a robust estimator on the inliers predicted with evaluated methods. The essential matrix is decomposed into a rotation matrix and a translation vector to compare with the ground truth.
We evaluate correspondence pruning in dynamic scenarios. The dynamic scenarios indicate there exist multiple consistencies between an image pair. We use a challenging dataset MULTI~\cite{multiConsist} for evaluation, whose samples have at least two consistencies. This dataset provides coordinates of inliers, and the putative correspondences without ground truth. We generates labels for putative correspondences by searching in putative inliers with KDtree~\cite{ramasubramanian1992fast}.

\noindent\textbf{Metrics.} We compute Area Under the Curve (AUC) to evaluate the pose accuracy at thresholds $5^{\circ}$, $10^{\circ}$, and $20^{\circ}$, which is also the metric used by~\cite{superglue,lmcnet} in camera pose estimation. We compute the precision, recall, and F1-scores with a symmetric epipolar distance threshold $\sigma=10^{-4}$ in camera pose estimation and with the computed labels in dynamic scenes.

\subsection{Camera Pose Estimation}

\noindent\textbf{Baselines.} We consider LMF~\cite{lmcnet}, CNe~\cite{cne}, ACNe~\cite{acne}, OANet~\cite{oanet}, and LMCNet~\cite{lmcnet} as  baselines and include the results of classical correspondence pruning approaches~\cite{magsac,ma2019locality,gms,lin2017code}. 
Learnable matchers, such as SuperGlue~\cite{superglue}, act on an extended putative set compared to sparse correspondence filters. We thus do not include them as the baselines. 
For implementation, we follow official released pretained models on YFCC100M with SIFT and our model is trained with the same setting. Considering the generalization performance, we evaluate all methods with the fixed models in different settings, \textit{e.g.}, other different datasets and various descriptors.

\begin{table}[t]\Huge
       \centering
       \resizebox{0.59\linewidth}{!}{
        \begin{tabular}{@{}lccc@{}}
        \toprule
        Method & \makebox[0.15\textwidth]{Precision} & \makebox[0.12\textwidth]{Recall} & F1-Scores\\
        \midrule
        CNe    &66.53  &35.67 &43.98 \\
        ACNe      &68.82  &57.57 &61.00 \\
        OANet        &67.80  &55.63 &59.18 \\
        LMCNet   & 65.08 & 53.08 & 56.00\\
        \network{}  &\textbf{71.40}  &\textbf{69.66} &\textbf{68.34} \\
        \bottomrule
       \end{tabular}
       }
    \caption{Results on MULTI. Best performance is in boldface.}
    \label{tab:multi}
\end{table}
    
\begin{table}[t]\huge
    \centering
    \resizebox{0.42\linewidth}{!}{
    \begin{tabular}{@{}lcc@{}}
    \toprule
    {Method} & \makebox[0.15\textwidth]{YFCC.} & SUN.\\
    \midrule
    
    CNe      &18.03  &{5.34}\\
    ACNe     &19.42  &5.05 \\
    OANet        &{19.83}  &4.75\\
    LMCNet      &{16.90}  &{1.12}\\
    \network{}        &\textbf{20.75}  &\textbf{5.66}\\
    \bottomrule
  \end{tabular}
  }
  \caption{Generalization with SuperPoint~(AUC@${5^{\circ}}$).}
    \label{tab:sp-ransac}
\end{table}

\begin{figure}[!t]
  \centering

    \includegraphics[width=0.46\textwidth]{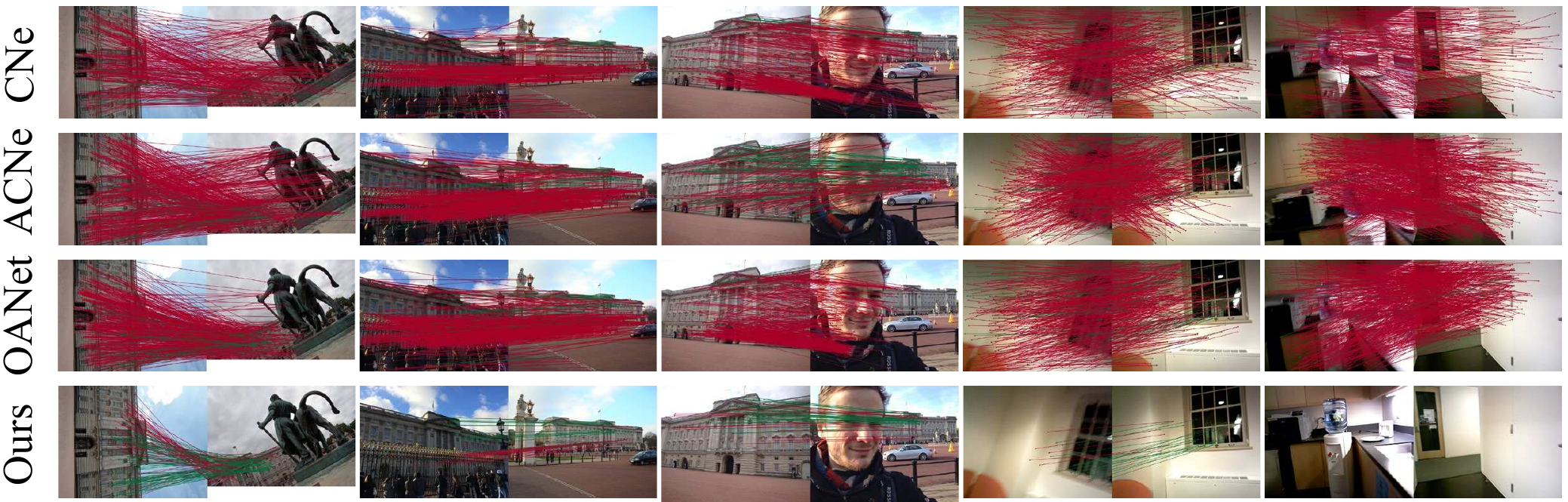}
    
  \caption{Visualizations of camera pose estimation. Red/Green indicates incorrect/correct correspondences. }
  \label{fig:vis}
\end{figure}
\noindent\textbf{Results.} Quantitative results are shown in \cref{tab:sift}. We observe that, \network{} reports the second best performance (a comparable performance against the best performance) in YFCC100M and the best performance in SUN3D. Further, to explore the reason in performance differences, we calculate the F-scores with different inlier ratios. As shown in \cref{fig:short}, \network{} outperforms baselines with $+1.7\%$ average improvement ($0\sim20\%$ inlier ratios and $80\sim100\%$ outlier ratios) on YFCC100M and $+8.5\%$ average improvement on SUN3D. It indicates the better pruning behavior of \network{} in high-outlier-ratio cases. \cref{fig:vis} shows the visual results with high outlier ratios. \network{} still correctly identifies inliers, while compared baselines identify a mass of outliers. Interestingly, as shown in the last column, \network{} can reject all correspondences when they are irregular.

We further evaluate the generalization ability in different settings and report results in \cref{tab:sp-ransac}. \network{} achieves leading performances in all settings, which indicates a clear advantage in generalization. Results on ScanNet\cite{dai2017scannet} can be found in the supplementary. A similar tendency can be observed. We  measure  the  inference  time  for 2048 correspondences over $4000$ runs and report the average time in~\cref{tab:parm}. We further take the time that generates putative correspondences (SIFT$\&$NN) as reference time. Despite achieving superior performance, LMCNet takes more than $1.4$ times as long as generating correspondences. Per to~\cref{tab:sift},~\cref{tab:sp-ransac}, and~\cref{tab:parm}, ANA-Net reports an approaching performance against LMCNet while running $14$ times faster.

\subsection{The MULTI Dataset}
Since MULTI~\cite{multiConsist} only contains 45 image pairs, models trained in camera pose estimation are used here. In practice, we test them with the same form in camera pose estimation: predicting inlier probability from putative correspondences. We report results of precision, recall, and F1-scores in \cref{tab:multi}. In MULTI, \network{} exhibits the ability to capture diverse consistencies, implied by higher precision, recall, and F-scores. Contrary to the strict correspondence pruning in camera pose estimation, in this task \network{} is tolerant to consistent correspondences (even with different consistencies) and therefore could preserve more correspondences compared with other baselines. 

\subsection{Ablation Study} Here we conduct ablation studies to understand how our method works and which part contributes to the most of performance improvement. We first compare different formulations discussed in 
``Second-Order Attention''. Results in \cref{tab:ab on hform}(Left) suggest our approximations are effective and efficient. Our approximations reduce the complexity of modeling second-order context from $\mathcal{O}(N^3)$ to $\mathcal{O}(N)$ without much performance loss. We remark that the slightly better performance of the quadratic form over the cubic one is due to the optimization difficulty of the naive implementation. For the second-order attention, we measure with different N's and results are shown in~\cref{tab:ab on hform}(Right). While the performance of the quadratic form seems better than the linear one, we recommend to use the linear form in practice, because the cost of its use almost comes for free.

\begin{table}[t]\Large
    \centering
    \resizebox{0.85\linewidth}{!}{
    \begin{tabular}{@{}lccc|ccc@{}}
    \toprule
     Form   & AUC@${5^{\circ}}$& @${10^{\circ}}$ & @${20^{\circ}}$     & N=2048     & 4096 & 8192\\
    \midrule
    Cub. &20.91 &38.62 &56.71 & +5.35 & +93.83 &+914.15  \\
    Qua. &21.28  &38.82 &56.86 & +4.72 & +7.41   &+26.85 \\
    Lin. &20.75 &38.11 &56.08 & +4.07 &+4.70     &+8.89 \\
    \bottomrule
    \end{tabular}
    }
 \caption{Left: Evaluation on different forms of second-order attention.We report the AUC on YFCC100M. Right: Times(ms). We measure extra(``+'') inference time(ms) introduced by second-order attention.} 
     \label{tab:ab on hform}
\end{table}
\begin{table}[t]\Huge
  \centering
     \resizebox{0.87\linewidth}{!}{
    \begin{tabular}{@{}ccc|cc|ccc@{}}
    \toprule
    Method& SOC & LNN & Parm.(k) & +Parm.&AUC@${5^{\circ}}$ & + step &+ total\\
    \midrule
    CNe &   -   &   -  &    397 &   -  &  18.03     &   - &\multirow{2}{*}{+1.39}\\
    ACNe&   -   &   -   &   410 &   +13&  19.42   & +1.39&\\ \hline
    \multirow{3}{*}{Ours}    &       &       &   524 &   +124&  19.73   &+0.31&\multirow{3}{*}{+1.33}    \\
        &\checkmark&    &   662 &   +138&  20.57   &+0.84&     \\      
    &\checkmark&\checkmark& $\approx$662&   +0.02  &20.75   &+0.18&    \\
    \bottomrule
\end{tabular}
}
      \caption{Ablation studies on YFCC100M. `LNN' indicates the learnable non-linear normalization. ``+'' is a comparison with the previous line.}
         \label{tab:ablation}
\end{table}

With the linear form, we further decompose the ANA-Block and report results in \cref{tab:ablation}. The baseline model only introduces the first-order attentive context, which  can  be  viewed  as  an  extension  of ACNe, maintains better performance against ACNe.
And all our components contribute to performance improvement. Specifically, the baseline brings a $+0.31$ improvement. The second-order attention (with vanilla $\tt sigmoid(\cdot)$ for non-linear normalization) brings a $+0.8$ improvement. It should be noted that its parameters increase is brought by the necessary encoder after the SOC layer as described in ``attention-consistent context enhancement'', 
 which dose not increase the capacity of the baseline. Besides, the proposed learnable non-linear normalization brings additional $+0.2$ improvement with only $20$ extra parameters. Together, considering ACNe is the follow-up work of CNe, it improves CNe by $1.39$. Instead our implemented first-order attention outperforms ACNe by $+0.31$  and the inclusion of the second-order attention improves ACNe by $1.33$.

\begin{figure}[t]
\centering
    \includegraphics[width=0.45\textwidth]{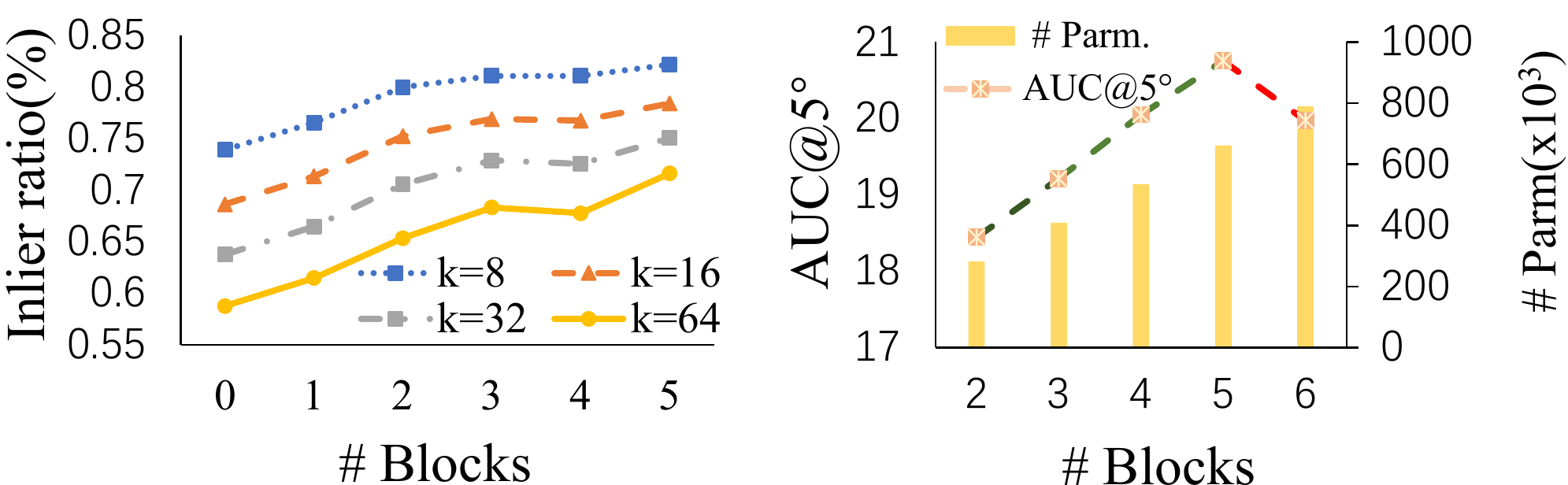}
  \caption{Left: Inlier ratios under different $k$'s with changing $\#$~blocks. 
    We collect intermediate features with different number of ANA blocks. 
    The features are used to calculate inlier ratios in the neighbors of correct correspondences with different neighbor size $k$.
  Right: AUC and $\#$~parameters with changing $\#$~blocks.}
     \label{fig:knnresults-a}
\end{figure}

\noindent\textbf{How the ANA-Block works.} We collect $k-$nearest neighbors using KNN for inliers from each stacked blocks with different $k$ and calculate the average inlier ratio. The number of blocks $L=5$. \cref{fig:knnresults-a} (Left) shows that, when more blocks are stacked, inliers tend to find additional inliers from their neighbors. The clustered inliers bodes well for subsequent inliers identification. 
It also embodies a inlier-clustering trend. Finally, we evaluate the performance with different $L$'s and analyze the number of parameters. As shown in \cref{fig:knnresults-a}(Right), the repetitive use of ANA-Block yields an increase in AUC, while the AUC slightly drops when $L=6$. Perhaps the distinction between inliers and outliers cannot be further enlarged with
available information such that performance is saturated. 

\begin{figure}[t]
\centering
    \includegraphics[width=0.40\textwidth]{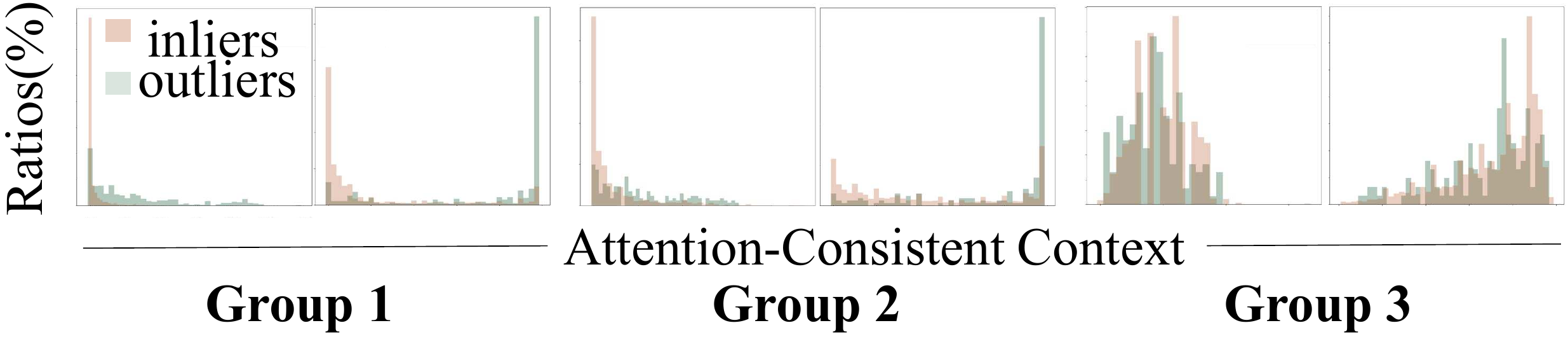}
    \caption{Conditions of applying attention-consistent context. Group 1: a clear bimodal distribution; Group 2: a weak bimodal distribution; Group 3: a confusing distribution.} 
    \label{fig:at sm visual}
\end{figure}

\noindent\textbf{How the second-order attention works.} Different from the conventional context considered in segmentation, attention-consistent context reflects a soft number of neighbors of attention weights, where inliers should own more neighbors, while outliers have less. This type of context
can be visualized directly.
As shown in \cref{fig:at sm visual}, there are $3$ typical situations when attention-consistent context is applied (each group visualizes an unnormalized and a normalized attention-consistent context): the group $1$ shows an ideal situation with a bimodal distribution between inliers and outliers; the bimodal shape is slightly degraded in the group $2$; and the group $3$ shows a confusing case where the context may not be helpful. The confusing case can be caused by high outlier ratios. We remark that, the multi-head architecture proposed by transformer can somewhat alleviate this confusing case, because it is not likely that the confusing case appear in each head at the same time. 

\section{Conclusion}
We introduce the idea of attention in attention to model second-order attention. We present a naive formulation and two approximated formulations to capture attention-consistent context. We show that attention-consistent context can be complementary with feature-consistent context to address correspondence pruning problems effectively and efficiently. We believe the combination of the first- and second-order attention points out a new direction for the attention mechanism and shows potential to extend to additional tasks where attention mechanism matters. 

 \section*{Acknowledgements}
This work was supported in part by the National Natural Science Foundation of China~(Grant No.U$1913602$) and  by Huawei Technologies CO., LTD.~(Grant No.YBN$2020085023$).
\bibliography{aaai23}

\end{document}